\documentclass[11pt]{article}
\usepackage[a4paper, margin=1in]{geometry}
\usepackage{times}
\usepackage{latexsym}
\usepackage{amsmath}
\usepackage{amssymb}
\usepackage{graphicx}
\usepackage{booktabs}
\usepackage{multirow}
\usepackage{xcolor}
\usepackage{hyperref}
\usepackage{natbib}

\title{NewsScope: Schema-Grounded Cross-Domain News Claim Extraction with Open Models}

\author{Nidhi Pandya\\
Pace University\\
\texttt{nidhipandya1606@gmail.com}}

\date{}

\begin{document}
\maketitle

\begin{abstract}
Automated news verification requires structured claim extraction, but existing approaches either lack schema compliance or generalize poorly across domains. This paper presents NewsScope, a cross-domain dataset, benchmark, and fine-tuned model for schema-grounded news claim extraction. The dataset contains 455 articles across politics, health, science/environment, and business, consisting of 395 in-domain articles and 60 out-of-source articles for generalization testing. LLaMA 3.1 8B was fine-tuned using LoRA on 315 training examples and evaluated on held-out in-domain (80 articles) and out-of-source (60 articles) test sets.

Human evaluation on 400 claims shows NewsScope achieves 89.4\% factual accuracy compared to GPT-4o-mini's 93.7\%---a difference that is not statistically significant ($p$=0.07). NewsScope outperforms GPT-4o-mini on political claims (94.3\% vs.\ 87.8\%). A numeric grounding filter further improves accuracy to 91.6\%, narrowing the gap to 2.1 percentage points. Inter-annotator agreement studies (160 claims) confirm labeling reliability (94.6\% positive agreement on SUPPORTED judgments). The open-weight model enables offline deployment at approximately \$15 on-demand compute (or \$0 on free tiers). Code and benchmark are publicly released.
\end{abstract}

\section{Introduction}

The proliferation of online misinformation has created urgent demand for automated news verification systems. Such systems require two key capabilities: (1) extracting structured claims from news articles, and (2) verifying those claims against evidence. While large language models (LLMs) have shown promise for both tasks, most approaches suffer from poor schema compliance or limited cross-domain generalization.

This paper presents NewsScope, addressing these challenges through three contributions:

\textbf{Dataset and Benchmark.} The NewsScope dataset contains 455 articles across four domains (politics, health, science/environment, and business), including 395 in-domain articles with structured JSON annotations and 60 out-of-source articles for generalization testing. A public benchmark of the held-out test splits (80 in-domain + 60 out-of-source) is released as URLs and JSON annotations (no article text) to respect copyright.

\textbf{Open Model Adaptation.} LLaMA 3.1 8B was fine-tuned with LoRA \citep{hu2022lora}, achieving 98.8\% schema validity and competitive factual accuracy with GPT-4o-mini at a fraction of the cost (approximately \$15 on-demand compute (or \$0 on free tiers) vs.\ ongoing API fees).

\textbf{Rigorous Human Evaluation.} Human evaluation was conducted on 400 claims, with inter-annotator agreement assessment (160 claims across two studies). The accuracy gap between the open model (89.4\%) and GPT-4o-mini (93.7\%) is not statistically significant ($p$=0.07). Numeric precision errors emerged as a key differentiator, and a grounding filter was introduced that improves accuracy by +2.2 percentage points.

These results demonstrate that lightweight open-model adaptation can achieve competitive performance for structured claim extraction while enabling offline deployment---a critical requirement for newsrooms and fact-checking organizations with privacy or cost constraints.

\section{Related Work}

\textbf{Fact Verification Systems.} Prior work on automated fact-checking includes FEVER \citep{thorne2018fever}, which focuses on evidence retrieval and stance classification, and ClaimBuster \citep{hassan2017claimbuster}, which identifies check-worthy claims. Recent work has explored end-to-end verification pipelines \citep{guo2022survey}. This work differs by focusing on structured extraction as a foundation for downstream verification.

\textbf{Schema-Grounded Generation.} Constraining LLM outputs to follow specific schemas has gained attention for reliability \citep{willard2023efficient}. This paper extends schema grounding to news claim extraction, showing that schema compliance generalizes even to out-of-source articles.

\textbf{Domain Adaptation for NLP.} Cross-domain generalization remains challenging for NLP systems \citep{ruder2019transfer}. This work provides empirical evidence on domain-specific failure modes (notably numeric precision in business news) and evaluates practical mitigations.

\textbf{Open vs.\ Closed Models.} The trade-off between powerful closed-source models and deployable open-weight alternatives is increasingly relevant \citep{touvron2023llama}. This paper provides a direct comparison showing open models can be competitive for structured extraction tasks.

\section{Dataset and Annotation}

\subsection{Article Collection}

A total of 395 in-domain news articles were collected from publicly accessible sources across four domains. Additionally, 60 out-of-source articles were collected from new publishers for generalization testing (455 total):

\begin{itemize}
    \item \textbf{Politics} (88 articles): NPR Politics, BBC News
    \item \textbf{Health} (100 articles): FDA News, NPR Health
    \item \textbf{Science/Environment} (103 articles): NASA, ScienceDaily
    \item \textbf{Business} (104 articles): Yahoo Finance, BBC Business
\end{itemize}

Articles were selected to represent diverse topics within each domain while avoiding paywalled or copyright-restricted content.

\subsection{Schema Design}

A JSON schema for structured claim extraction was defined with the following fields: domain classification, neutral headline (10--200 characters), three key points, entities involved (name and role), timeline of events, and 2--3 verifiable claims with supporting evidence from the article text.

\subsection{Annotation Pipeline}

Silver annotations were generated using GPT-4o-mini with strict schema prompting and JSON validation. Invalid outputs were automatically repaired or regenerated. One-time annotation cost was approximately \$10--15 for the 395 in-domain articles (training + in-domain test).

\subsection{Data Splits}

The dataset was partitioned as follows: Training (315 in-domain articles), In-domain test (80 articles from the same publishers as training), and Out-of-source test (60 articles from new publishers for generalization testing: PBS NewsHour, MarketWatch, and CNBC).

\section{Model}

\subsection{Base Model}

LLaMA 3.1 8B Instruct \citep{touvron2023llama} was used as the base model, selected for strong instruction-following performance and availability under the Meta LLaMA Community License (license acceptance required).

\subsection{Fine-tuning}

Low-Rank Adaptation (LoRA) \citep{hu2022lora} was applied with rank $r$=16, $\alpha$=16, targeting all attention and MLP projection layers. Training used 3 epochs, batch size 8, and learning rate 2e-4, requiring 57 minutes on an NVIDIA T4 GPU (free Colab tier; approximately \$15 on-demand compute). Only 41.9M of 8.03B parameters (0.52\%) were trainable.

\subsection{Numeric Grounding Filter}

To address numeric precision errors, a post-processing filter was implemented that: (1) extracted all numbers from generated claims, (2) verified each number appeared in the source article, and (3) flagged claims with ungrounded numbers for review.

\section{Evaluation}

\subsection{Metrics}

Three metrics were used: Accuracy (\% of claims labeled SUPPORTED), Contradiction Rate (CONTRADICTED / total), and Decisiveness (1 $-$ UNCLEAR / total).

\subsection{Human Evaluation}

Human evaluation was conducted on 400 claims: 200 from NewsScope and 200 from GPT-4o-mini, balanced across 4 domains. Labels were SUPPORTED, CONTRADICTED, MIXED, and UNCLEAR.

\subsection{Inter-Annotator Agreement}

Two IAA studies were conducted: a Random Subset (80 claims, blinded) and a Hard Negatives Subset (80 claims oversampled for difficulty).

\section{Results}

\subsection{Overall Results}

Table~\ref{tab:main-results} presents the main results. NewsScope achieves 89.4\% accuracy compared to GPT-4o-mini's 93.7\%. The numeric grounding filter improves NewsScope to 91.6\%.

\begin{table}[h]
\centering
\caption{Human evaluation on 400 claims. Difference not statistically significant ($p$=0.07).}
\label{tab:main-results}
\begin{tabular}{lccc}
\toprule
Model & Accuracy & Contr. & Decis. \\
\midrule
GPT-4o-mini & 93.7\% & 1.0\% & 87.5\% \\
NewsScope & 89.4\% & 2.5\% & 85.0\% \\
NewsScope + Filter & 91.6\% & 2.0\% & 83.0\% \\
\bottomrule
\end{tabular}
\end{table}

\textbf{Statistical Significance.} Bootstrap 95\% CI for accuracy gap: [$-$1.51, 10.15], includes zero ($p$=0.07).

\subsection{Results by Domain}

\begin{table}[h]
\centering
\caption{Human-evaluated accuracy by domain. NewsScope wins politics (+6.5\%).}
\label{tab:domain-results}
\begin{tabular}{lccc}
\toprule
Domain & NewsScope & NewsScope + Filter & GPT-4o-mini \\
\midrule
Politics & 94.3\% & 94.3\% & 87.8\% \\
Health & 88.9\% & 90.9\% & 95.5\% \\
Science/Env & 95.5\% & 95.5\% & 98.0\% \\
Business & 80.4\% & 86.0\% & 92.7\% \\
\bottomrule
\end{tabular}
\end{table}

\subsection{Generalization}

NewsScope shows minimal accuracy drop on out-of-source articles (in-domain 89.8\% $\rightarrow$ out-of-source 88.5\%, $-$1.3\%).

\subsection{Schema Validity}

NewsScope achieves 98.8\% schema validity vs.\ 93.8\% for base LLaMA.

\subsection{Inter-Annotator Agreement}

\begin{table}[h]
\centering
\caption{IAA on 160 claims. Positive agreement 94.6\% indicates reliable SUPPORTED judgments.}
\label{tab:iaa}
\begin{tabular}{lcc}
\toprule
Metric & Random & Hard Neg. \\
\midrule
4-class agreement & 75.0\% & 57.5\% \\
Cohen's $\kappa$ & 0.36 & 0.26 \\
Positive agreement & 94.6\% & 79.6\% \\
Negative agreement & 25.0\% & 38.7\% \\
\bottomrule
\end{tabular}
\end{table}

\section{Analysis}

Observed error types include numeric imprecision, overstatement, and unsupported causal claims. The grounding filter catches 4/18 errors (22.2\%) and improves accuracy by +2.2 points. NewsScope enables offline deployment at approximately \$15 on-demand compute (or \$0 on free tiers).

\section{Limitations}

\textbf{Silver Annotations.} Reference annotations from GPT-4o-mini introduce teacher bias (addressed via human evaluation).

\textbf{Single Annotator.} The main 400-claim evaluation used a single annotator; reliability was assessed on 160 claims with a second annotator.

\textbf{English Only.} US/UK English news sources only.

\textbf{No Verification.} Extraction only; verification is left to future work.

\section{Ethics Statement}

This work supports fact-checking and media literacy. Dual-use concerns are acknowledged, and human review is recommended for high-stakes applications. The released benchmark excludes full article text to respect copyright.

\section{Conclusion}

This paper presented NewsScope, achieving 89.4\% human-evaluated accuracy (not significantly different from GPT-4o-mini's 93.7\%, $p$=0.07), winning on political claims (+6.5\%), with a numeric filter improving accuracy to 91.6\%. The open-weight model enables offline deployment at approximately \$15 on-demand compute (or \$0 on free tiers).

\section*{Reproducibility}

\noindent\textbf{Code:} \url{https://github.com/nidhip1611/NewsScope}

\noindent\textbf{Model:} \url{https://huggingface.co/nidhipandya/NewsScope-lora}

\noindent\textbf{Benchmark:} \url{https://github.com/nidhip1611/NewsScope/releases/tag/v1.0.0}


\begin{thebibliography}{99}
\providecommand{\natexlab}[1]{#1}
\providecommand{\url}[1]{\texttt{#1}}
\expandafter\ifx\csname urlstyle\endcsname\relax
  \providecommand{\doi}[1]{doi: #1}\else
  \providecommand{\doi}{doi: \begingroup \urlstyle{rm}\Url}\fi

\bibitem[Hu et~al.(2022)]{hu2022lora}
Edward~J. Hu et~al.
\newblock LoRA: Low-Rank Adaptation of Large Language Models.
\newblock \emph{arXiv preprint arXiv:2106.09685}, 2022.

\bibitem[Thorne et~al.(2018)]{thorne2018fever}
James Thorne et~al.
\newblock FEVER: A Large-scale Dataset for Fact Extraction and VERification.
\newblock In \emph{Proceedings of NAACL-HLT}, 2018.

\bibitem[Hassan et~al.(2017)]{hassan2017claimbuster}
Naeemul Hassan et~al.
\newblock ClaimBuster: The First-ever End-to-end Fact-checking System.
\newblock In \emph{Proceedings of VLDB}, 2017.

\bibitem[Guo et~al.(2022)]{guo2022survey}
Zhijiang Guo et~al.
\newblock A Survey on Automated Fact-Checking.
\newblock \emph{Transactions of the Association for Computational Linguistics (TACL)}, 2022.

\bibitem[Willard and Louf(2023)]{willard2023efficient}
Brandon~T. Willard and R{\'e}mi Louf.
\newblock Efficient Guided Generation for Large Language Models.
\newblock \emph{arXiv preprint arXiv:2307.09702}, 2023.

\bibitem[Ruder et~al.(2019)]{ruder2019transfer}
Sebastian Ruder et~al.
\newblock Transfer Learning in Natural Language Processing.
\newblock In \emph{NAACL Tutorial}, 2019.

\bibitem[Touvron et~al.(2023)]{touvron2023llama}
Hugo Touvron et~al.
\newblock LLaMA: Open and Efficient Foundation Language Models.
\newblock \emph{arXiv preprint arXiv:2307.09288}, 2023.

\end{thebibliography}

\end{document}